\documentclass[runningheads]{llncs}
\usepackage{graphicx}
\usepackage{times}
\usepackage{latexsym}
\usepackage{tcolorbox}
\usepackage{url}
\usepackage{array,siunitx}
\newcolumntype{M}{>{\centering\arraybackslash}m{1cm}}
\newcolumntype{G}{>{\centering\arraybackslash}m{1.15cm}}
\newcolumntype{H}{>{\centering\arraybackslash}m{1.1cm}}
\usepackage{multirow}
\usepackage{tcolorbox}
\usepackage{hyperref}
\usepackage{todonotes}

\begin{document}
\title{How to Pre-Train Your Model? \newline
Comparison of Different Pre-Training Models for Biomedical Question Answering}

\titlerunning{Comparison of Different Pre-Training Models for Biomedical Question Answering}

\author{Sanjay Kamath\inst{1,3}\and
Brigitte Grau\inst{1,2}\and
Yue Ma\inst{3}}

\authorrunning{Kamath et al.}
\institute{LIMSI, CNRS, Universit\'{e} Paris-Saclay, Orsay, France \and
 ENSIIE, Universit\'{e} Paris-Saclay, \'{E}vry, France \\ 
 \and
LRI, Univ. Paris-Sud, CNRS, Universit\'{e} Paris-Saclay, Orsay, France \\ \email{\{sanjay, ma\}@lri.fr, bg@limsi.fr}}

\maketitle              

\begin{abstract}
Using deep learning models on small scale datasets would result in overfitting. To overcome this problem, the process of pre-training a model and fine-tuning it to the small scale dataset has been used extensively in domains such as image processing. 
Similarly for question answering, pre-training and fine-tuning can be done in several ways. Commonly reading comprehension models are used for pre-training, but we show that other types of pre-training can work better.
We compare two pre-training models based on reading comprehension and open domain question answering models and determine the performance when fine-tuned and tested over BIOASQ question answering dataset. We find open domain question answering model to be a better fit for this task rather than reading comprehension model.

\keywords{Deep learning  \and Reading comprehension \and Open domain question answering}
\end{abstract}

\section{Introduction}

Deep learning models have been widely used in several NLP tasks since the emergence of large scale labelled datasets. In Question Answering (QA) specifically on open domain, several neural network models have been introduced, such as Convolutional Neural Networks(CNN), Recurrent Neural Networks(RNN) using GRUs or LSTMs and attention mechanisms, Self-attention networks (Transformers), and Pretrained language models like ELMO, BERT which can be fine-tuned to Question Answering task.
Several kinds of Question Answering (QA) related tasks are widely studied such as \textit{Answer Sentence Selection}, \textit{Reading Comprehension} and \textit{Open QA}. 

\textit{Reading Comprehension (RC)} is a QA task where a question and a relevant paragraph are given and the goal is to extract the answer string present in the paragraph. The main assumption of this task is that the answer is present in the paragraph, like in SQUAD v1.0 \cite{rajpurkar2016squad}. Variants of this task include unanswerable questions such as in \cite{rajpurkar2018know}. Answers are usually short phrases or entities. There is a leaderboard on SQUAD dataset\footnote{https://rajpurkar.github.io/SQuAD-explorer/} which showcases lot of models built for this task \cite{seo2016bidirectional,chen2017reading,bert}.

\textit{Open QA} is a QA task where a question is given and the goal is to retrieve an answer. An answer has to be retrieved from a set of documents or passages of textual sources as Wikipedia articles or news. Answers are also usually short phrases or entities. 
In NN approaches for Open QA, generally answers are extracted using a reading comprehension model on the subset of the retrieved documents or passages considered as relevant \cite{DBLP:journals/corr/DhingraMC17,joshi2017triviaqa}.

One of the main differences between \textit{Reading Comprehension (RC)} and \textit{Open QA} tasks is that the answer must be present in the paragraphs (or documents) for Reading Comprehension, but for Open QA this condition might not hold true because the retrieved documents considered to be relevant to the question might not contain the answer.  Another characteristic is that in the Open QA task, several paragraphs or documents contain the answer.

The BIOASQ Phase B task provides dataset for biomedical question answering which is a small scale labelled dataset for factoid questions (779 question in BIOASQ 7). Each question is associated with multiple relevant paragraphs, some irrelevant ones, and one or several answers. 
The work of \cite{wiese-etal-2017-neural} transforms the BIOASQ Phase B dataset into the format of a Reading Comprehension task where each question has an answer text along with the offset in a paragraph which contains the answer. If a paragraph does not contain an answer, it is discarded. This modification of the BIOASQ dataset enables to use a RC model off-the-shelf. 

By using such a model on BIOASQ dataset which is a small scale labelled dataset, it  will not result in similar performance as on the large scale open domain datasets due to overfitting. One way of overcoming this problem as reported by \cite{cicling,wiese-etal-2017-neural} is by pre-training a deep learning model on a large scale dataset and fine-tuning the same model to the target small scale dataset. The intuition is that the model learns better representations when learnt on a large scale dataset than having a randomly initialized model trained only on the small scale dataset.


However the BIOASQ task resembles more towards an \textit{Open QA} task than a \textit{RC} task because of the existence of paragraphs without answers even though they are considered relevant. Thus we propose a new way to tackle the BIOASQ task by using an Open QA model that takes into account this particularity.

We present a comparison of using different pre-training models (reading comprehension and open QA models) for BIOASQ question answering task and also report the performance of a single model without pre-training and without fine-tuning to show the importance of this process. 

We report the performance of our model on different datasets and show that in some cases it outperforms the state-of-the-art systems of BIOASQ \cite{wiese-etal-2017-neural,kamath-etal-2018-adaption,DBLP:journals/corr/abs-1901-08746} in average.

\section{State of The Art}

Since BIOASQ 5, deep learning methods were introduced by \cite{wiese-etal-2017-neural} by automatically adapting BIOASQ QA task as \textit{Reading Comprehension} task and pre-training the model with SQUAD v1.0 dataset. Similar approach of pre-training and fine-tuning are used by \cite{kamath-etal-2018-adaption} who pre-train their models using DRQA \cite{chen2017reading} and BioBert \cite{DBLP:journals/corr/abs-1901-08746} using Bert \cite{bert}.

The models discussed in this article for the BIOASQ task use automatic annotations as done by \cite{wiese2017neural} who transform the BIOASQ dataset into reading comprehension dataset by using the gold standard answer strings by searching them in the snippets for exact match and are treated as answers if only they are found in the snippets, i.e., the answer string must be a substring of the snippet. 

As the NN models participating to BIOASQ are based on domain adaptation, we first present general approaches used for that purpose before presenting how it is done by the BIOASQ NN models. 

\subsection{Domain Adaptation}

Pre-training is a training process started from randomly initialized model weights. 
Fine-tuning is also a training process but started from the model weights of pre-trained model and not randomly initialized model weights.
Both pre-training and fine-tuning together can be termed as \textit{Domain Adaptation} when the domain of data used for pre-training and fine-tuning are different. For example, open domain and biomedical domain.

Pre-training and fine-tuning or domain adaptation can be done in several ways.
The general approaches are listed below.

   \textit{Type 1}  - The task remains the same for pre-training and fine-tuning. Pre-training should be done on  a large scale dataset from random initialization of parameters. Fine-tuning should be done on a small scale dataset by loading the model parameters from pre-trained model rather than random initialization.
     This approach is used when a target dataset is small scaled and using it to train a deep neural network would result in overfitting. This type of pre-training is common in computer vision field where models are pre-trained on Imagenet \cite{Russakovsky2015} and fine-tuned on target image classification datasets.
    
   \textit{Type 2} - The tasks are different for pre-training and fine-tuning. Pre-training should be done on a large scale dataset from random initialization of parameters. Fine-tuning should be done on a different model which uses certain parameters from the pre-trained model which are frozen (non-trainable) and learns some parameters which are randomly initialized on a different task.
   These approaches in NLP were initially proposed for sequence labelling tasks by \cite{peters2017semi} which were later evolved into ELMO (Embedding Language Models) by \cite{peters2018deep} which significantly improved the state of the art across a broad range of challenging NLP tasks such as question answering, textual entailment and sentiment analysis.
This type of method uses special contextual text embeddings obtained from the pre-trained models that are added as features into downstream models built for another task. 

    \textit{Type 3}-The tasks are different for pre-training and fine-tuning. Pre-training should be done on a large scale dataset from random initialization of parameters. Fine-tuning should be done on the pre-trained model by modifying certain layers to fit to the new task. Newly added layers can be randomly initialised and pre-trained model layers together with newly added ones are trained on the new task.
 This approach is similar to \textit{Type 2} approach with a difference that the reference model can be slightly modified for target task rather than building a different model. This type of approach proposed by \cite{bert} is being widely used in NLP tasks such as question answering, textual entailment, sentiment analysis, named entity recognition, relation extraction etc. which are easily done by modifying a final output layer of the original model and fine-tuned. 
Fine-tuning can be done either by learning the whole model parameters or learning only a part of the model by freezing the rest. 

Our work uses \textit{Type 1} domain adaptation and compares the results with \textit{Type 3} domain adaptation results reported by \cite{DBLP:journals/corr/abs-1901-08746}.

\subsection{Deep learning and Domain Adaptation in BIOASQ}

The work by \cite{wiese2017neural} comes under \textit{Type 1} domain adaptation using SQUAD v1.0 pre-training. 
Since the introduction of pre-trained language models by \cite{peters2017semi}, works by \cite{peters2018deep,bert} have been used in several NLP tasks and have been proven to outperform many prior state of the art models. 
BioBert by \cite{DBLP:journals/corr/abs-1901-08746}
have been shown to be useful in biomedical domain tasks such as named entity recognition, relation extraction and BIOASQ question answering. This work belongs to \textit{Type 3} domain adaptation methods where the authors use Bert model by \cite{bert} and re-train it on the same task but on biomedical domain texts. Later this model is modified for different biomedical tasks and tested. This method, as reported in the paper \cite{DBLP:journals/corr/abs-1901-08746}, fetches state of the art scores on BIOASQ QA task which is listed in Table \ref{scores} under \textit{BioBert} column.

\section{Question Answering Tasks and Models}

In this section, we describe the two kinds of question answering tasks and the related models we used for domain adaptation towards biomedical domain. 

\subsection{Tasks}
Question Answering (QA) is a field of research which lies in the intersection of Natural Language Processing and Information Retrieval disciplines. Several types of tasks exists which are commonly referred as Question Answering. 
In this article, we focus on Reading Comprehension (RC) task a.k.a Machine Reading task, and Open Domain Question Answering a.k.a Open QA or Open Question Answering. 

Reading Comprehension task contains questions, a relevant paragraph and answers from the paragraph. RC is also called as Answer Extraction because the answer is known to be present in the paragraph. 

Open QA task contains questions and their short answers without any paragraphs. Open QA can be formulated as a parent task which involves two child tasks, 1) Retrieving the relevant paragraphs for a question and 2) Extracting a short answer from the paragraphs. In Open QA, the first task is generally referred as paragraph selection or answer sentence selection
and the second task is often modelled as Reading Comprehension although there exists several relevant and irrelevant paragraphs. Open QA models should distinguish if the paragraph is relevant and then extract the answer unlike the RC models. 

BIOASQ phase B task is a question answering task with questions in biomedical domain. For a question, there are relevant documents, paragraphs, answers given. 
Below is an example from the dataset.
\begin{small}
\begin{tcolorbox}
\textbf{Q: Which calcium channels does ethosuximide target?}\\
\textbf{A: T-type calcium channels}\\
\textbf{P1: \textcolor{green}{..neuropathic pain is blocked by ethosuximide, known to block T-type calcium channels,..}}\\
\textbf{P2: \textcolor{red}{Theta rhythms remained disrupted during a subsequent week of withdrawal but were restored with the T-type channel blocker ethosuximide.}}
\end{tcolorbox}
\end{small}

The goal of BIOASQ question answering task is to extract the correct answer from supporting data. As shown in the example, one paragraph (P1) has the gold standard answer and the other (P2) does not (i.e. it does not contain the exact match of the answer string). Therefore this resembles more like an \textit{Open QA} task than a \textit{Reading Comprehension} task. 

\begin{center}
\begin{figure}[h]
\centering
\includegraphics[width=0.99\textwidth]{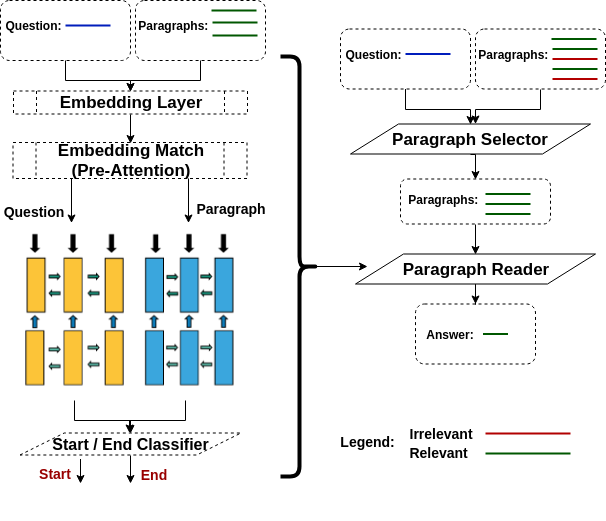}
\caption{Left: DRQA - Paragraph Reader  (\textit{RC task}). Right: PSPR - Paragraph Selector and Paragraph Reader model (\textit{Open QA task}) } \label{drqa_pic}
\end{figure}
\end{center}

The evolution of deep learning methods led to the emergence of large scale datasets for these two types of tasks in open domain Question Answering.
We use two models for \textit{RC} and \textit{Open QA} which are shown in the Figure \ref{drqa_pic} and are described below.

\subsection{Reading Comprehension -  DRQA model}
\textit{DRQA}'s document reader developed by \cite{chen2017reading} is a simple LSTM model for \textit{Reading Comprehension} task which takes as input a question and a paragraph and aims at extracting an answer from the paragraph. 
As per the assumption of the \textit{RC task}, the answer is always present inside the paragraph as a substring.
An overview of the model can be seen on the left figure of figure \ref{drqa_pic}. 
Both the question and the paragraph are tokenized and their word embeddings are used for the model 
i.e. question words $Q = \{q_{1},.....,q_{m}\}$ and paragraph words  $S = \{s_{1},.....,s_{n}\}$ are sequences which are encoded using an embedding layer of dimension $D$.

\begin{equation}
E(Q) = \{E(q_{1}),..,E(q_{m})\}
\end{equation}
\begin{equation}
E(S) = \{E(s_{1}),..,E(s_{n})\}
\end{equation}

A pre-attention mechanism captures the similarity between paragraph words and question words in the same layer. For this purpose, a feature $\mathcal{F}align$ shown in Equation \ref{eq1} is added as a feature to the LSTM layer. 

\begin{equation}
\mathcal{F}align(p_{i}) = \Sigma_{j} a_{i,j} E(q_j)
\label{eq1}
\end{equation}
Where $a_{i,j}$ is, 
\begin{equation}
a_{i,j} = \frac{exp \,(\alpha(E(s_{i})) \cdot \alpha(E(q_{j}))}{ \Sigma_{j'} \ exp(\alpha(E(s_{i})) \cdot \alpha(E(q_{j'}))}
\end{equation}

which computes the dot product between nonlinear mappings of word embeddings of question and paragraph.

They are followed by a 3-layer Bidirectional LSTM layers for both question and sentence encodings.
\begin{equation}
\{E(q_{1}),..,E(q_{n})\} = \textnormal{Bi-LSTM}(\{\tilde{E}(q_{1}),..,\tilde{E}(q_{n}\})
\end{equation}
\begin{equation}
\{E(s_{1}),..,E(s_{n})\} = \textnormal{Bi-LSTM}(\{\tilde{E}(s_{1}),..,\tilde{E}(s_{n}\})
\end{equation}

These LSTM states are connected to two independent classifiers that use a bilinear term to capture the similarity between paragraph words and question words and compute the probabilities of each token being start or end of the answer span. 

\begin{equation}
\label{eq5}
    P_{\text {start}}(i) \propto \exp \left(\mathbf{p}_{i} \mathbf{W}_{s} \mathbf{q}\right)
\end{equation}
\begin{equation}
\label{eq6}
    P_{\text {end}}(i) \propto \exp \left(\mathbf{p}_{i} \mathbf{W}_{e} \mathbf{q}\right)
\end{equation}

During prediction, we choose the best span from token $i$ to token $i^{\prime}$ such that $i$ $\leq$ $i^{\prime}$ $\leq$ $i+15$ and $P_{start}(i)$ $\times$ $P_{end}$$(i^\prime)$ is maximized.

To make the scores compatible across paragraphs in one or several retrieved documents, unnormalized exponential is used and an argmax is taken over all considered paragraph spans for the final predictions which are offset to start and end of an answer span in the paragraph.

Answer probability for the answer span can be computed among several other answer spans as shown in section \ref{openqamodel} in equation \ref{eq44}. This probability score can be used to return top 5 answers for BIOASQ task which is explained in section \ref{shortcontext}.

\subsection{Open QA  - PSPR model}
\label{openqamodel}
\textit{PSPR} model is an Open QA model by \cite{lin2018denoising} whose overview is presented in the right figure of Figure \ref{drqa_pic}. 
This model has two parts, namely Paragraph Selector (PS) and Paragraph Reader (PR) in a cascade fashion. 
Although \textit{PSPR} model contains a \textit{Reading Comprehension} task submodule for answer extraction, the main difference comes from learning the answer extraction module using the paragraph probabilities computed by the paragraph selector.
Paragraphs for the questions are retrieved using an information retrieval technique. Then the Paragraph Selector model predicts a probability distribution $Pr\left(p_{i} | q, P\right)$ over all the retrieved paragraphs where $P$ is the set of paragraphs for the question. 

The Paragraph Reader model extracts answer spans as shown in the \textit{DRQA} model and predicts a probability $Pr\left(a | q, p_{i}\right)$ for each answer span where $p_{i}$ is $i^{th}$ paragraph in Paragraph set $P$. 

The reader model gives two probabilities (one for start and one for end token given by two classifiers) as described in equation \ref{eq5} and \ref{eq6}.
The answer probability $Pr(a | q, P)$ is computed as shown below:

\begin{equation}
\label{eq44}
    Pr\left(a | q, p_{i}\right)=\sum_{j} Pr\left(a_{s}^{j}\right) Pr\left(a_{e}^{j}\right)
\end{equation}

The answer with highest probability is returned as the final prediction.

The Paragraph Selector uses tokenized question words $Q = \{q_{1},.....,q_{m}\}$  and tokenized paragraph words $P = \{p_{1},.....,p_{n}\}$  which are encoded using an embedding layer of dimension $D$.

\begin{equation}
E(Q) = \{E(q_{1}),..,E(q_{m})\} 
\end{equation}
\begin{equation}
E(P) = \{E(p_{1}),..,E(p_{n})\}
\end{equation}

A RNN layer encodes the contextual information of the sequence. 
\begin{equation}
\{E(q_{1}),..,E(q_{m})\} = \textnormal{RNN}(\{\tilde{E}(q_{1}),..,\tilde{E}(q_{m}\})
\end{equation}
\begin{equation}
\{E(p_{1}),..,E(p_{n})\} = \textnormal{RNN}(\{\tilde{E}(p_{1}),..,\tilde{E}(p_{n}\})
\end{equation}

Using this hidden representation, a self attention operation is applied to get the question representation $q$:

\begin{equation}
    \hat{\mathbf{q}}=\sum_{j} \alpha^{j} \hat{\mathbf{q}}^{j}
\end{equation}

where $\alpha^{j}$ encodes the importance of each question word against the other question words which is calculated as:
\begin{equation}
    \alpha_{i}=\frac{\exp \left(\mathbf{w}_{b} \mathbf{q}_{i}\right)}{\sum_{j} \exp \left(\mathbf{w} b \mathbf{q}_{j}\right)}
\end{equation}
Where $w$ is the learnt weight vector.
Finally, the probability of each paragraph is calculated via a max-pooling and a softmax layer as shown below:
\begin{equation}
\label{eq3}
    Pr\left(p_{i} | q, P\right)=softmax\left(\max _{j}\left(\hat{\mathbf{p}}_{i}^{j} \mathbf{W} \mathbf{q}\right)\right)
\end{equation}
where $\textbf{W}$ is a learnt weight matrix.

Since not all the paragraphs contain an answer in the Open QA setting, the probability scores from equation \ref{eq3} should indicate if there exists an answer or not.

While training, the paragraphs containing the answer are highlighted as 1 and the rest as 0. And while testing, the paragraph with highest probability is chosen to extract the answer. 
Combining the two probabilities, the overall answer is chosen by choosing the highest probable answer from $Pr(a | q, P)$ for a question $q$ which is calculated as :

\begin{equation}
\label{eq2}
Pr(a | q, P)=\sum_{p_{i} \in P} Pr\left(a | q, p_{i}\right) Pr\left(p_{i} | q, P\right)
\end{equation}

For Paragraph Reader, the \textit{DRQA} model can be used directly as shown in the figure \ref{drqa_pic} with small differences.
During training the reader model extracts the answer only when there is an answer in it. i.e. when the paragraph probability score (equation \ref{eq3}) of that paragraph is 1. 
During testing, the reader model extracts the answer from the paragraph which has the highest probability score. 

Only difference while adapting this to BIOASQ is that number of answers to be extracted for BIOASQ is top 5 and not 1. Because of this, instead of choosing from only the top most probable paragraph, we select top 5 answers from combined probability scores in equation \ref{eq2}, which might consider 1 or several paragraphs to extract answers from.



\subsection{Domain adaptation for BioAsq task}

Pre-training and fine-tuning or domain adaptation can be done in several ways, the following is a general abstraction of the two processes:

\begin{itemize}
    \item \textbf{Type 1} \begin{itemize}
        \item Build a reference model (Model R) for task A.
        \item Train the Model R with a sufficiently large scale dataset on task A.
        \item Save the Model R weights and model. 
        \item DO NOT randomly initialize Model R weights, instead use the saved weights of model R from the previous step.
        \item Train the Model R again on a small scale target dataset on task A. 
        \item Predict and Evaluate.
        \end{itemize}
     \item \textbf{Type 2} \begin{itemize}
        \item Build a reference model (Model R) for task A.
        \item Train the Model R with a sufficiently large scale dataset on task A.
        \item Save the Model R weights and model. 
        \item Build a model from scratch (Model S) for task B.
        \item Model S is built to use some features from model R.
        \item Initialize the model S weights randomly (apart from the Model R features which are used right away). 
        \item Train the Model S on a large or small scale target dataset on task B. 
        \item Predict and Evaluate.
        \end{itemize}
   \item \textbf{Type 3} \begin{itemize}
    \item Build a reference model (Model R) for task A.
    \item Train the Model R with a sufficiently large scale dataset on task A.
    \item Save the Model R weights and model. 
    \item Build a Model S, which is built upon Model R (by Adding or Modifying the input or output layers of Model R to work on task B).
    \item Initialize the model S weights randomly (All of Model R features can be used right away, or some partially used and partially randomized for training). 
    \item Train the Model S on a large or small scale target dataset on task B. 
    \item Predict and Evaluate.
    \end{itemize} 
\end{itemize}

In this work, we apply \textit{Type 1} domain adaptation.

\subsubsection{The data for pre-training}
Two datasets correspond to each of the two tasks: SQUAD V1.0 dataset for \textit{RC} task and QUASAR-T dataset for \textit{Open QA} task and we show below their differences.

\begin{itemize}
    \item QUASAR-T which is based on Trivia questions is 
    generated synthetically, and SQUAD is annotated manually by humans on a crowd-sourcing platform. 
    \item Each question in QUASAR-T is associated to 100 sentence-level passages retrieved from ClueWeb09 dataset based on Lucene, whereas SQUAD 1.0 has 1 relevant paragraph. 
    \item Some paragraphs in QUASAR-T do not have an answer. \footnote{SQUAD 2.0 is a variant of SQUAD dataset which contains questions without answers. We do not use this because the reference models also do not use v2.0 to pre-train.}
\end{itemize}

Comparing the above differences with BIOASQ dataset, the QUASAR-T dataset resembles more closely to BIOASQ than that of SQUAD v1.0 due to the following reasons.
\begin{itemize}
    \item BIOASQ data has more than 1 relevant paragraphs per question.
    \item Some paragraphs do not have an answer.
\end{itemize}

\section{Experiments and Results}
\label{results}

For fine-tuning the models we use BIOASQ datasets. The statistics of different sets are reported in the Table \ref{stats}.

\begin{table}[h!]
\begin{center}

\begin{tabular}{ | c | c | c | c |  }
\hline
 Datasets & Train & Dev & Test    \\ \hline\hline
\multirow{1}{*}{BIOASQ 4b}
 & 427 & 59 & 161 \\ \hline

\multirow{1}{*}{BIOASQ 5b}
&544 & 75 & 150  \\ \hline

\multirow{1}{*}{BIOASQ 6b}
& 685 & 94 & 161   \\ \hline 

\multirow{1}{*}{SQUAD v1.0}
& 87,599 & 10,570 & 9,533   \\ \hline 

\multirow{1}{*}{QUASAR-T}
& 37,012 & 3,000& 3,000    \\ \hline 
\end{tabular}
\caption{Datasets used in the experiments along with their splits. The numbers represent number of questions.}
\label{stats}
\end{center}
\end{table}

\begin{table}[h!]
\begin{center}

\begin{tabular}{ | c | c | c | c | c |  }
\hline
 Datasets & Metrics   & No-Pre & No-Fine &  Pre+Fine     \\ \hline\hline
\multirow{4}{*}{4b}
 & S.Acc &08.98 & 23.96 & \textbf{24.00} \\ 
& L.Acc &16.56 & 35.26 & \textbf{39.21}   \\ 
 & MRR &11.36 & 28.40 & \textbf{29.34}  \\ \hline

\multirow{4}{*}{5b}
& S.Acc &25.91 &32.17 & \textbf{32.43}   \\ 
& L.Acc  &34.86 & 45.58& \textbf{47.73}   \\ 
& MRR & 29.27 & 36.75 & \textbf{38.37}  \\ \hline

\multirow{4}{*}{6b}
& S.Acc &13.40 &26.24 & \textbf{26.72 }   \\ 
& L.Acc &27.08 &40.60 & \textbf{43.72}    \\ 
& MRR &19.20 &32.57 & \textbf{33.80}     \\ \hline \hline

\multirow{4}{*}{Average}
& S.Acc &16.09 & 27.45 & \textbf{27.71}  \\ 
& L.Acc  & 26.16&40.48 & \textbf{43.55}     \\ 
& MRR &19.94 &32.57 & \textbf{33.83}  \\ \hline 

\end{tabular}
\caption{Results reporting the importance of Pre-Training and Fine-Tuning a model. DRQA model by \cite{chen2017reading} is used for these experiments. S.Acc is Strict Accuracy, L.Acc is Lenient Accuracy (the correct answer is in the top 5) and MRR is Mean Reciprocal Rank.}
\label{nopre}
\end{center}
\end{table}

\subsection{Importance of \textit{Pre-Training} and \textit{Fine-Tuning}  }
To show the importance of \textit{Pre-Training} and \textit{Fine-Tuning} for domain adaptation to biomedical domain, we experimented three approaches on a single model \textit{DRQA} without altering any hyperparameters. (Default parameters as used by \cite{chen2017reading}\footnote{https://github.com/facebookresearch/DrQA})

1) \textit{No-Pre} model is the \textit{DRQA} model trained on BIOASQ dataset only. 

2) \textit{No-Fine} model is the \textit{DRQA} model trained on SQUAD v1.0 dataset only. 

3) \textit{Pre+Fine} is the \textit{DRQA} model trained on SQUAD v1.0 dataset and fine-tuned on BIOASQ dataset. 

Results are shown in the Table \ref{nopre} for different BIOASQ test sets. 
When a model is only trained on a small dataset like BIOASQ, the results are very low as shown in the column \textit{No-Pre}.
When a model is trained on a large  dataset like SQUAD v1.0, the model can be straight away used to predict results on the biomedical dataset. \textit{No-Fine} shows an improvement doing so, against \textit{No-Pre}.
Lastly, \textit{Pre+Fine} pre-training on a large scale dataset and fine-tuning on the biomedical dataset shows an improvement over the other approaches.
This set of experiments show that the best approach is to do pre-training and fine-tuning on smaller domain specific datasets.

\subsection{Experiments with the two QA modellings: DRQA and PSPR}
\label{shortcontext}
\begin{table}[h!]
\begin{center}

\begin{tabular}{ | c | c |  c | c | c | c | }
\hline
 Datasets & Metrics  & BioBert by \cite{DBLP:journals/corr/abs-1901-08746} &  DRQA  &  DRQA+PS  &PSPR   \\ \hline\hline
\multirow{4}{*}{4b}
 & S.Acc & \textbf{36.48} &24.00 &26.22  &   30.28   \\ 
& L.Acc& \textbf{48.89} &39.21 & 32.33& 40.34 \\ 
 & MRR & \textbf{41.05}&29.34 &26.54  & 34.19  \\ \hline

\multirow{4}{*}{5b}
& S.Acc &41.56&32.43  & 30.62 &   \textbf{46.59} \\ 
& L.Acc  & \textbf{54.00}&47.73   &47.86 &   53.76  \\ 
& MRR &46.32&38.37  &36.96 &  \textbf{49.55} \\ \hline

\multirow{4}{*}{6b}
& S.Acc &35.58&26.72  & 26.50 &  \textbf{43.91}  \\ 
& L.Acc &\textbf{51.39}&43.72  & 42.16  &   51.34  \\ 
& MRR &42.51&33.80  & 32.07  &  \textbf{45.70}  \\ \hline \hline

\multirow{4}{*}{Average}
& S.Acc &37.87& 27.71  & 27.78 &   \textbf{40.26}  \\ 
& L.Acc  & \textbf{51.43} & 43.55   &40.78 &  48.48 \\ 
& MRR & \textbf{43.29}& 33.83   &31.85 &  43.14  \\ \hline 

\end{tabular}

\caption{ \textit{DRQA} is the Reading Comprehension model by \cite{chen2017reading}, \textit{PSPR} is the Open QA model by \cite{lin2018denoising}, \textit{DRQA+PS} is answers chosen with scores by multiplying answer probabilities of DRQA with Paragraph Selector probabilities of PSPR. SOTA scores are reported by \cite{DBLP:journals/corr/abs-1901-08746} who average the best scores from each batch (possibly from multiple different models). Results from BIOASQ 4b, 5b and 6b test sets. 7b test set cannot be evaluated yet due to lack of gold standard answers. S.Acc is Strict Accuracy, L.Acc is Lenient Accuracy and MRR is Mean Reciprocal Rank. Experiments are done with the original BIOASQ data.}
\label{scores}
\end{center}
\end{table}

In this section, we experiment mainly with the two models for a \textit{Reading Comprehension} task and an \textit{Open QA} task. 

For studying the modelling of the BIOASQ QA task as a \textit{Reading Comprehension} task, we use SQUAD v1.0 dataset for pre-training and experiment with the \textit{DRQA} model. For studying its modelling as an \textit{Open QA} task, we use QUASAR-T dataset for pre-training and experiment with \textit{PSPR} model.

We also experiment using the paragraph probabilities for reranking the DRQA answers and choosing the top 5. 
As the \textit{PSPR} model is a cascaded model with paragraph selector and paragraph reader, we use the paragraph probabilities predicted by the paragraph selector and multiply them with the answer probabilities obtained using \textit{DRQA} model to select the top 5 answers which have combined higher probabilities. 
Note that this approach is not the same as cascaded \textit{PSPR} model because the \textit{PSPR} model's reader model uses the paragraph probabilities to learn the extraction of answers which might have a different impact.

For obtaining top 5 answers with \textit{DRQA} model, if there are more than 5 paragraphs for the question, we take 1 answer from each paragraph and choose top 5 based on the answer probabilities. We do this to make sure each paragraph's top answer is contributed towards top 5 answers.
If there are less than 5 paragraphs, we take top 5 answers based on answer probabilities to keep it simple.


Results are shown in Table \ref{scores} for different BIOASQ test sets. We compare different model results with BioBert scores reported in \cite{DBLP:journals/corr/abs-1901-08746}. The scores from \textit{PSPR} model shows that the performance is better than BioBert on Strict and Lenient accuracy on 5b and 6b test sets. By taking paragraph probability into account \textit{PSPR} allows to better rank top 1 correct answer than BioBert which extracts answers only from the longer pre-processed paragraphs which have correct answers.

Although \textit{PSPR} has a reader model similar to \textit{DRQA}, considering the paragraph probability seems to improve the answer extraction in \textit{PSPR} model.


\subsection{Experiments with longer contexts (modified BIOASQ data)}
\begin{table}[h!]
\begin{center}

\begin{tabular}{ | c | c |  c | c | c | c |  }
\hline
 Datasets & Metrics  & SOTA  & DRQA & BioBert-Unaltered & BioBert by \cite{DBLP:journals/corr/abs-1901-08746}   \\ \hline\hline
\multirow{4}{*}{4b}
 & S.Acc &20.59 & 18.49&13.08 & \textbf{36.48}   \\ 
& L.Acc& 29.24 & 32.51 &18.54 &  \textbf{48.89}  \\ 
 & MRR &24.04&23.88 & 15.48 &\textbf{41.05}    \\ \hline

\multirow{4}{*}{5b}
& S.Acc & \textbf{41.82}& 28.92 & 22.84&  41.56 \\ 
& L.Acc  &57.43& 46.54 &32.46 & \textbf{54.00}   \\ 
& MRR & \textbf{47.73}& 35.88 & 25.94 &46.32 \\ \hline

\multirow{4}{*}{6b}
& S.Acc &25.12& 21.70 &16.35 &\textbf{35.58}  \\ 
& L.Acc &40.20& 41.51 & 22.61& \textbf{51.39}  \\ 
& MRR &29.28&28.60 &18.72 & \textbf{42.51}  \\ \hline \hline

\multirow{4}{*}{Average}
& S.Acc &29.18& 23.03 & 17.42 &  \textbf{37.87}  \\ 
& L.Acc  &42.29 & 40.18 & 24.53 &\textbf{51.43} \\ 
& MRR &33.68& 29.45 & 20.04 &\textbf{43.29} \\ \hline 

\end{tabular}

\caption{Experiments with data containing longer contexts (Document level) by
\cite{DBLP:journals/corr/abs-1901-08746}. \newline  \textit{DRQA} is a Reading Comprehension model by \cite{chen2017reading}. \textit{BioBert-Unaltered} is the original BIOASQ dataset with questions and paragraphs which contain answers. \textit{Biobert by \cite{DBLP:journals/corr/abs-1901-08746}} is the modified BIOASQ dataset where the paragraphs are longer paragraphs (documents from respective articles), where all the models are pre-trained on SQUAD v1.0 dataset and finetuned on BIOASQ dataset. SOTA scores are reported by \cite{DBLP:journals/corr/abs-1901-08746} who average the best scores from each batch (possibly from multiple different models). Results from BIOASQ 4b, 5b and 6b test sets. 7b test set cannot be evaluated yet due to lack of gold standard answers. S.Acc is Strict Accuracy, L.Acc is Lenient Accuracy and MRR is Mean Reciprocal Rank.}
\label{scores_ALTER}
\end{center}
\end{table}

For the BIOASQ task we noted that the method used by \cite{DBLP:journals/corr/abs-1901-08746} with BioBert modifies the original paragraphs. For computing the BioBert model, the authors have retrained the original Bert model by \cite{bert}, using Pubmed and PMC articles. For applying it on the BIOASQ task, the authors use longer documents (instead of the actual snippets) from Pubmed corresponding to the data given by BIOASQ in the "documents" field to access the Pubmed documents for each question. 
Therefore the modification of the dataset leads to different results for BioBert compared to the performance on the regular BIOASQ dataset. The exact pre-processing of BIOASQ dataset in order to do this is not very clear from the paper, however the authors release the modified dataset in their repository\footnote{https://github.com/dmis-lab/biobert}.

In order to evaluate the importance of this data modification, we did three experiments:
1) DRQA with longer contexts 2) BioBert with unaltered data from BIOASQ 3) BioBert results with modified paragraphs and as reported by \cite{DBLP:journals/corr/abs-1901-08746} in their paper.

The results are shown in  Table \ref{scores_ALTER}.
The results of \textit{BioBert} is as presented in \cite{DBLP:journals/corr/abs-1901-08746} where the authors have fine-tuned the models first using SQUAD v1.0 dataset and adapted it to BIOASQ data. 
 We use the modified dataset to experiment it with the \textit{DRQA} model to determine if it would improve the performance of the pre+fine \textit{DRQA} model as reported in Table \ref{nopre}. We got lower performances to that of the \textit{DRQA} model trained on the original BIOASQ data.

For comparison, we try the BioBert model on the original BIOASQ data i.e. paragraphs given by BIOASQ data and not pre-processed. The results in Table \ref{scores_ALTER}, under the column \textit{BioBert-Unaltered} represents these results. It is evident that the modification performed on the BIOASQ data fetches better results using BioBert model.

\section{Conclusion}

In this work we have shown the importance of pre-training and fine-tuning process a.k.a domain adaptation for biomedical domain question answering. We have also compared two QA models based on i.e. 1) Reading Comprehension task 2) Open QA task, and found that the performance is better when using an Open QA model than a Reading Comprehension model. 

Based on a different pre-processing done by \cite{DBLP:journals/corr/abs-1901-08746} on the biomedical dataset by using longer contexts from documents than shorter contexts, we found that the Reading Comprehension model performs worse on the pre-processed longer contexts compared to the shorter contexts originally given by BIOASQ data. 

On the other end, a large pre-trained language model such as BERT performs much better on the pre-processed longer contexts than shorter contexts. 
Future work shall focus on training BERT model on Open QA task which will better suit the BIOASQ dataset. 

\section{Acknowledgements}
This work is funded by the ANR project GoAsQ (ANR-15-CE23-0022).

\bibliographystyle{splncs04}
\bibliography{samplepaper}%
\end{document}